# Zone-based Keyword Spotting in Bangla and Devanagari Documents

[a]Ayan Kumar Bhunia, [b]Partha Pratim Roy*, [c]Umapada Pal

[a]Dept. of ECE, Institute of Engineering & Management, Kolkata, India
[b]Dept. of CSE, Indian Institute of Technology Roorkee, India
[c]CVPR Unit, Indian Statistical Institute, Kolkata, India
[b]email: proy.fcs@iitr.ac.in, TEL: +91-1332-284816

## Abstract

In this paper we present a word spotting system in text lines for offline Indic scripts such as Bangla (Bengali) and Devanagari. Recently, it was shown that zone-wise recognition method improves the word recognition performance than conventional full word recognition system in Indic scripts [29]. Inspired with this idea we consider the zone segmentation approach and use middle zone information to improve the traditional word spotting performance. To avoid the problem of zone segmentation using heuristic approach, we propose here an HMM based approach to segment the upper and lower zone components from the text line images. The candidate keywords are searched from a line without segmenting characters or words. Also, we propose a novel feature combining foreground and background information of text line images for keyword-spotting by character filler models. A significant improvement in performance is noted by using both foreground and background information than their individual one. Pyramid Histogram of Oriented Gradient (PHOG) feature has been used in our word spotting framework. From the experiment, it has been noted that the proposed zone-segmentation based system outperforms traditional approaches of word spotting.

# 1. Introduction

Handwritten text recognition is one of the challenging problems in the field of pattern recognition. Due to the free-flow nature of handwriting and many writing variations, the recognition performance is not satisfactory even with sophisticated pre-processing and OCR techniques. While processing such handwritten documents, word spotting [3] techniques are useful to search the possible instances of specific/query words. For searching using "Word Spotting", it does not require OCR of the entire document. The presence of writing distortion does not create much problem in retrieving similar target words as these approaches do not involve recognition of either the characters of the query word or the query word itself. The features are extracted from the whole word and thus the methods try to find similar features in the target images. One of the drawbacks of these methods is that these require exact word segmentation prior to the matching. Word retrieval will not be feasible using feature matching in target image in the case of improper word segmentation. To tackle this problem, recently segmentation free methods [5, 10, 11] are proposed. Word spotting has been extensively studied [2, 4, 5] to detect a word in a handwritten document page (or line) as per the user's query keyword [5] or a template image [10, 13]. This way of searching or browsing approach often overcomes the problem of conventional recognition. Word spotting with Query by Example (QBE) principle takes instances of query word image for searching. Whereas Query by Text (QBT) [15] which uses learning based approach for retrieval proved more effective recently. Due to the success of "Word Spotting" with efficient technology such as Hidden Markov Model and Neural Network, it has been popular in extracting information from historical documents, handwritten forms, etc.

Classical approaches, proposed in Latin [5, 15], Arabic [2] scripts decompose text line into a sequence of vertical frames and features are extracted from each of them, and are fed to a decoding system to retrieve the text sequence of characters. Nevertheless, one of the bottlenecks of such systems is feature extraction. Especially, in Indian scripts (Bengali, Devanagari, etc.), a combination of vowels, modifiers, and characters lead to a huge number of character classes where recognition/spotting is still challenging [27]. The main problem of dealing with Bangla or Devanagari handwritten script is the free flow nature of handwriting and an immense amount of variation in writing styles. Besides that, there are many classes of characters and special upper and lower zone characters (i.e. modifiers) in Bangla and Devanagari scripts which make it even difficult to recognize than others (like English). In both Bangla and Devanagari scripts, there are approximately 50 basic characters including vowels and consonants [29]. The consonants may join with other consonants or vowels to form cluster and thus the number of characters in those two scripts is about 300. Upper and lower zone modifiers when get combined with the consonants form large set of combination, which have to considered as different character units for each combination separately, because the plain sliding window feature will need to capture the information in all zones for identifying the modifier properly. Recently, it was shown that zone-wise segmentation method improves the word recognition performance than conventional full word recognition system in Indic scripts [29]. Inspired with this idea we consider the zone segmentation approach and use middle zone information to improve the traditional word spotting performance. Different zones in Bangla and Devanagari words are shown in Fig.1. We show visually in Fig.2 the character class reduction using zone segmentation for both Devanagari and Bangla characters.

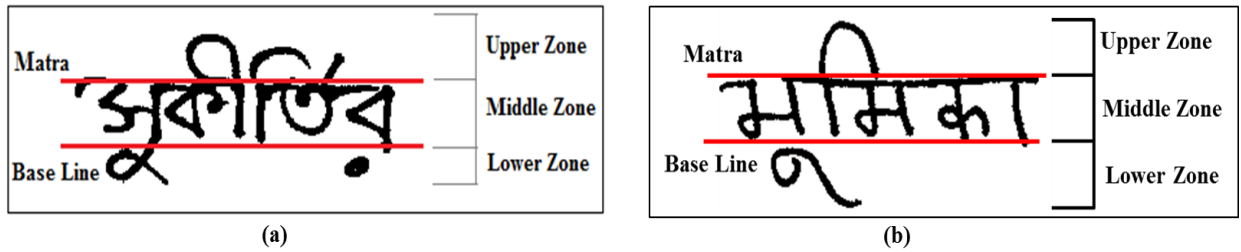

**Fig.1: Example showing presence of upper-middle-lower zone in (a) Bangla and (b) Devanagari scripts, respectively.**

**Fig.2: Example showing character units reduction using zone segmentation for Devanagari character 'क' and Bangla character 'ক' combining with different modifier combination.**

This paper presents a Query-by-Text based word spotting system in segmented text lines using Hidden Markov Models. We propose here a novel approach of combining foreground and background information of text images for keyword-spotting by character filler models. The candidate keywords are searched in line without segmenting character or words. A significant improvement in performance is noted by using both foreground and background information than each of one alone. We apply an HMM-based zone wise segmentation in text lines to boost the word spotting performance. The framework is applied in Indic scripts such as Bengali and Devanagari along with Latin script for evaluation. Proposed framework is shown graphically in Fig.3.

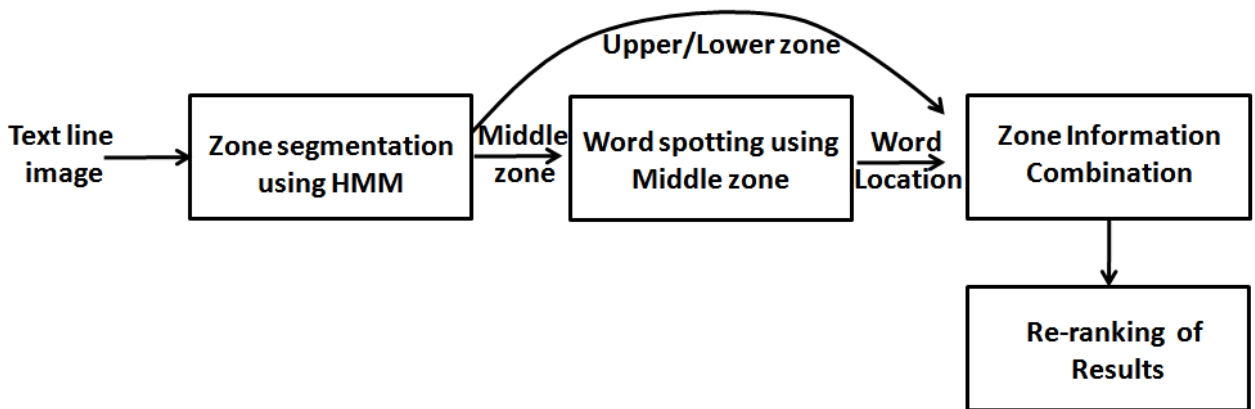

**Fig.3: Proposed framework of word spotting in Indic script**

## 1.1 Related Work

Handwritten word spotting is traditionally viewed as an image matching task between one or multiple query word images and a set of candidate word images in a database [3, 4]. There exist many pieces of work on word spotting in applications of postal documents [16, 17, 18], bank checks [19], digital libraries [20], historical documents [21], etc. A recent survey on recent word spotting methodologies on handwritten documents can be found in [32]. Different features have been explored by researchers. Corner features have been used by [22], GAF (gradient angular feature) was used by [23], etc.The techniques of query by example (QBE) or image template matching [10] was adopted by researchers in the early days of word spotting. In retrieval of important information from poorly written old documents [3], word spotting has been considered. Several local features have been used for achieving better performance among which some outperformed others in conjunction with Dynamic Time Wrapping (DTW). In [23] zones of interest (ZOI) part of the document have been extracted considering only the informative part to retrieve the query keyword. In [10] Scale- Invariant Feature Transform (SIFT) [25] features are extracted from similar size patches followed by transformation of feature space of the document to topic space using Latent Semantic Indexing. Then using cosine distance measure, patches with enough evidence are considered as spotted result.

The second approach of word spotting namely query by text (QBT) or the learning based approach [12, 13] which outperformed the older one, is being extensively used in recent systems. Here, word level segmentation has been avoided in [2, 26] using supervised learning model like HMM, BLSTM directly on the segmented text lines. HMM models are trained for each character and occurrence of specific character sequence is determined based on the probability scores which it returns for every testing text lines. Recurrent Neural Network (RNN) with Bidirectional Long Short-Term Memory (BLSTM) hidden nodes and Connectionist Temporal Classification (CTC) output layer has been explored for keyword spotting in [26]. Some work exists in which character template based [13] spotting has been considered whereas others depicts spotting at word level. Several works exist towards the application of word spotting such as keyword finding in historical documents [4, 13], searching and browsing through a digitized document, etc. A script independent word spotting method has also been proposed recently [2]. At each location of the window, a feature vector is extracted [15] and the sequence of feature vectors obtained in this fashion is modelled with Semi-continuous HMM. This approach has been tested only in segmented words.There exists several page level segmentation free techniques which uses scale invariant features (i.e. SIFT) [10]. A pyramidal histogram of character based training approach was proposed [33] for improved word retrieval performance.

Very few works have been done for keyword spotting in Indic script. Shape code based word-matching technique was used in Indic printed text lines [27]. Here,vertical shaped based feature, zonal information of extreme points, loop shape and position, crossing count and some background information has been used to search a query word.Segmentation-free method for spotting query word images in Bangla handwritten documents is done in [28] using Heat Kernel signature (HKS) to represent the local characteristics of detected key points.

The contributions of this paper are three folds. First we present a novel feature extraction method for word spotting using combination of foreground and background information. We noted that background information improves the word spotting performance significantly. Next, a zone-wise segmentation approach is used to reduce the number of character classes in Indic scripts. Using HMM training with middle-zone components, the performance improves. To boost the performance, a combination of zone information is used to reduce the false positives. Finally, we propose a HMM based zone alignment to improve the traditional projection based zone segmentation. The frame work for word spotting has been tested in Indic scripts namely Bangla and Devanagari. In depth experimental evaluation in a large dataset demonstrate the robustness of the proposed system. A preliminary experiment with foreground and background combination was presented in [31]. This paper extends it and present the details of word spotting in Indic script.

The rest of the paper is organized as follows. The word spotting framework is explained in details in Section 2. In Section 3, we discuss the zone segmentation approach using HMM and next word-spotting using middle zone information is explained. A combination of zone information is also presented in this section. We demonstrate the performance of our proposed framework in Section 4. Finally, conclusions and future work are presented in Section 5.

## 2. HMM-based Word Spotting Framework

The major goal which deals with word spotting is to detect specific keyword in a pool of document images. Our system is able to search arbitrary words in the text lines. For this purpose, the document image is first binarized with a global binarization method. Next, the binary document is segmented into individual text lines using a line segmentation algorithm [6]. For skew-correction; we consider all the points on the extreme bottom of the text stroke and use *Linear Regression* analysis on these points to find out the best fitted line. The slope of the straight line $\delta$ represents skew of the text. Thereafter, a rotation by $\delta$ is done to correct the skew [29]. We have determined the slant angle locally by dividing the text line images into segments. Then, the slant angle is determined and corrected using vertical projection histogram and Wigner Viller distribution [35]. Thus, slope and slant of the text line is normalized to cope up with different handwriting style. Fig.4. provides the graphical description of the word spotting framework where concatenated features are feed to HMM. Word spotting is being done using text line scoring based on the filler and character model of HMM. For the word spotting system we have used a novel feature extraction technique. Concatenation of foreground feature and background features are considered here. The details of each step are described below.

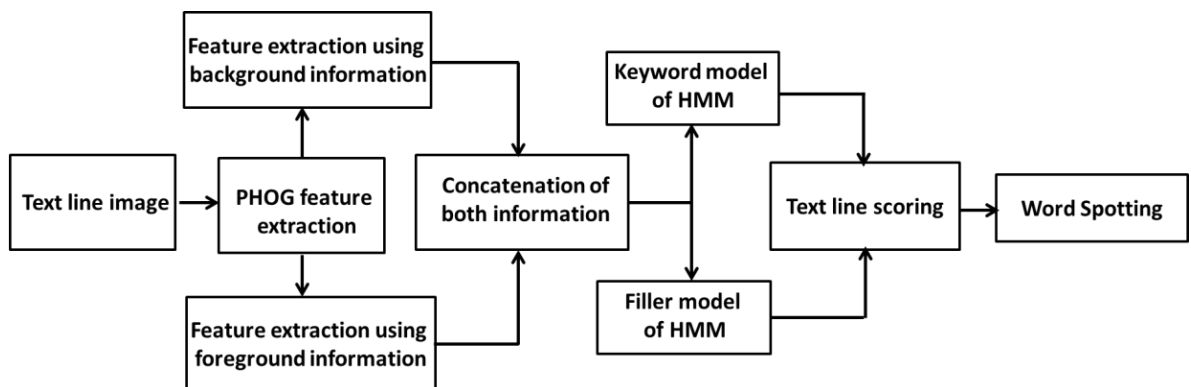

**Fig.4: HMM-based word spotting framework.**

**2.1. Feature extraction**

Feature is a representation of an image which is more descriptive than the image. PHOG feature has been found to provide better result in Bangla handwritten script recognition [7]. PHOG [12] is the spatial shape descriptor which gives the feature of the image by spatial layout and local shape, comprising of gradient orientation at each pyramid resolution level. To extract the feature from each sliding window, we have divided it into cells at several pyramid level. The grid has $4^N$ individual cells at N resolution level (i.e. N=0, 1, 2,..). Histogram of gradient orientation of each pixel is calculated from these individual cells and is quantized into L bins. Each bin indicates a particular octant in the angular radian space. The concatenation of all feature vectors at each pyramid resolution level gives 168 dimensional ($8 \times \sum_{i=0}^{N} 4^i = 168$) feature vectors considering 8 bins and limiting the level to N=2 in our implementation. The concatenation of all feature vectors at each pyramid resolution level gives 168 dimensional feature vectors considering 8 bins and limiting the level to N=2 in our implementation.

For calculating background information we take care of the morphology of character set in Bangla and Devanagari scripts. In Bangla or Devanagari script it is noted that most of the characters have a horizontal line (Shirorekha) at the upper part. When two or more characters sit side by side to form a word, the horizontal lines of the characters touch and generate a long line called head-line. Because of such touching nature the characters in a word create big white regions (spaces) in Bangla or Devanagari scripts. These empty spaces are found by water reservoir principle [11]. For each pair of joining characters we will get unique reservoir formation, these reservoirs contain information about the combination of characters forming the word. In Fig.5 the formation of bottom reservoirs are shown for Devanagari and Bangla text line, respectively.

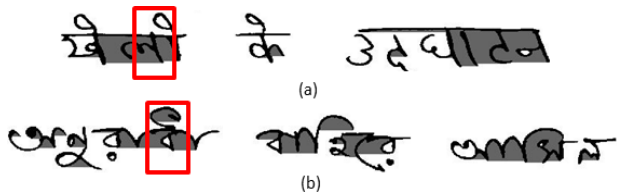

**Fig.5: Water Reservoir formation in (a) Devanagari and (b) Bangla text line image and position of sliding window is marked in red color.**

We have calculated PHOG feature from foreground as well as background regions, formed by the reservoir. These features are then concatenated for the final feature from the text line image. An illustration of feature extraction technique is given in Fig. 6.

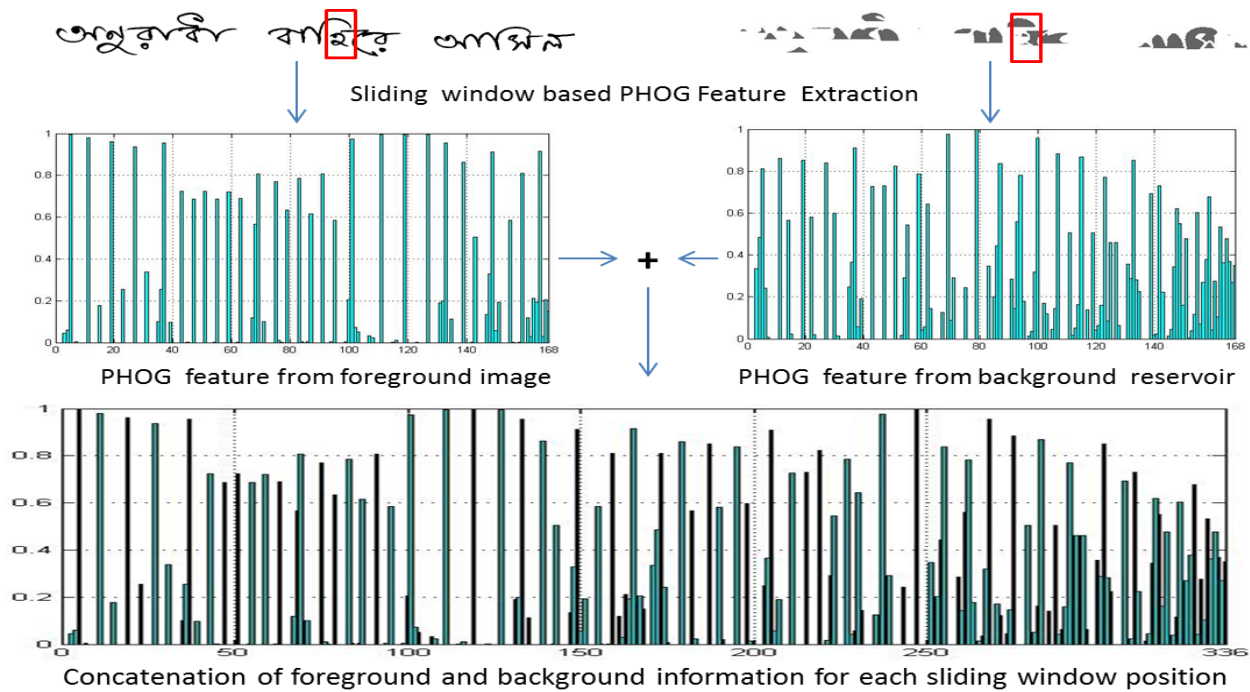

**Fig.6: Feature extraction method shown graphically. The features are extracted from the sliding window marked in red color.**

### 2.2. HMM-based Text line scoring

In the field of handwritten text recognition, Hidden Markov Models have been extensively used because of efficiency at recognition in the cases of touching characters, distorted characters even without being properly pre-processed [14]. The HMMs are capable of modelling cursive handwritten text which is difficult to segment for recognition.

The simplest model is the character HMM which consists of J hidden states ($S_1$, $S_2$ ... $S_J$) in a linear topology as an observation O where $i^{th}$ observation ($O_i$) represents an n-dimensional feature vector **x** modelled using a Gaussian Mixture Model (GMM) with probability $P_{S_j}(x)$, 1<j<J given by

$$P_{S_j}(x) = \sum_{k=1}^{G} W_{jk} N(x|\mu_{jk}, \Sigma_{jk})$$

where G is the number of Gaussians and $N$ refers to a multivariate Gaussian distribution with mean $\mu_{jk}$, covariance matrix $\Sigma_{jk}$ and probability $W_{jk}$ for $k^{th}$ Gaussian in state j**.**For training the model, firstly, sliding window based feature is extracted from labeled text line images with multiple words. The probability of the character model of the text line is then maximized by Baum-Welch algorithm assuming an initial output and transitional probabilities. Using the character HMM models, a filler model has been created which represents an arbitrary sequence of characters (Fig. 7(a)) Fig.7(b) shows the keyword model[5] which has been used in our system to spot a keyword in a text line image. The filler model represents a single character model consisting of any one of the characters among 'Char i's, where 1 ≤i≤N (see Fig.7(a)). A 'Space' model has been used in the keyword model shown in Fig.7(b) which is accounted for modeling white spaces.

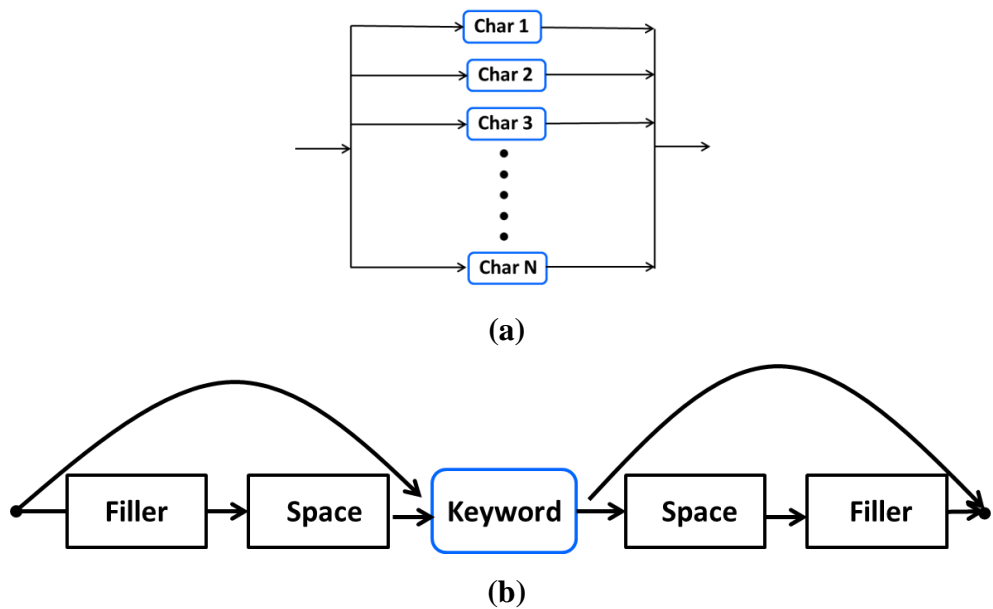

**Fig.7: Examples of (a) Filler Model and (b) Keyword Model**

Word spotting mechanism is based on the scoring [5] of text image(X) for the queried keyword (W). If the score value is greater than a certain threshold then it gives a positive value for the occurrence of that particular keyword in that text line. The score assigned to the text line image X for the keyword W is based on the posterior probability $P(W_j|X_{a,b})$ trained on keyword models where a and b correspond to starting and ending position of the keyword whereas $X_{a,b}$ gives the particular part of text line containing the keyword [5]. Applying Bayes' rule we get

$$\log p(W|X_{a,b}) = \log p(X_{a,b}|W) + \log p(W) - \log p(X_{a,b})$$

Considering equal probability we can ignore the term $\log p(W)$. The term $\log p(X_{a,b}|W)$ represents the keyword text line model (see Fig. 7(b)) where it is assumed that exact character sequence of the keyword to be present separated by 'Space'. The rest part of the text line is modelled with Filler text line model (see Fig. 7(a)). Then we can find the position a, b for the keyword alongside with the log-likelihood $\log p(X_{a,b}|W) = \log p(X_{a,b}|K)$.

$\log p(X_{a,b})$ is the unconstrained filler model F. The general conformance of the text image to the trained character models is given by obtained log-likelihood $\log p(X_{a,b}) = \log p(X_{a,b}|F)$. The difference between the log-likelihood value of keyword model and filler model is normalized with respect to the length of the word to get the final text line score.

$$Score(X, W) = \frac{[\log p(Xa, b | K) - \log p(Xa, b | F)]}{b - a}$$

Then this $Score(X, W)$ is compared with a certain threshold [5] for word spotting. This score is used for estimating the MAP (Mean Average Precision) value of our word spotting system.

| Query Keyword | Text line image | Result |
|---|---|---|
| সি পি এম | হাতে নন্দিল মেদিনীপুরে আবারও এক সি.পি.এম | ✓ |
| সি পি এম | জগদ্, মানবাজার সি.পি.এম সালবনি ব্লকে | ✓ |
| পুলিশ | পুলিশ ও নিরাপত্তা রক্ষীদের লক্ষ করিয়া | ✓ |
| পুলিশ | তখন ভীর পার্থিবি হুগুম্ হুগুম্ | ✗ |
| इसके | की गाथी भी इसके लिए | ✓ |
| इसके | इसके मद्देनजर धर्मतल्ला | ✓ |

Fig.8: Qualitative results of few word spotting where spotted word is indicated by red box and result is given by correct (by tick) and incorrect (by cross) labels.

## 3. Word Spotting using Zone Segmentation

As discussed earlier, full-zone wise character based HMM models may not be found to be effective in Indian scripts, especially in Bangla and Devanagari. Bangla and Devanagari scripts contain large combination of vowels, modifiers and characters which lead to a huge number of character classes. Hence, sufficient data for each class will be necessary for training the respective class models. To deal with this problem zone segmentation based word recognition approach has been adapted in [7, 29].This approach reduces the character classes drastically and makes it robust to model the character classes using lesser training data along with major improvement in recognition accuracy. The zone segmentation approach proposed in [29] will not be applicable as this system requires segmented words. Also, it involves recognition based segmentation for middle zone extraction. In this paper we include a HMM-based zone segmentation which does not require segmented word images. The zone segmentation is performed using Viterbi Forced alignment.

In literature, projection based analysis [24] has been used for segmenting upper and lower zones in printed text. But, zone segmentation in handwritten Indic script is challenging because of its inherent complex writing style and touching characters. To overcome this problem we propose here a novel HMM-based approach to segment middle zone from other zones. Given a text line image from the dataset, a sliding window of a fixed width is chosen to slide horizontally across the line image from left to right. Next, in each position of this window, a sub-window of fixed height is moved vertically from top to bottom. The image patches in each sub-window are trained using HMM for identifying three zones: upper, middle and lower zones. Hence, three different classes are used in HMM for training. This learning based zone segmentation method can take care the cases where the distance between consecutive words is not uniform or text lines with varying length, for example, in forms or tables where the lines are very short. A pictorial view of the framework is given in Fig.9.

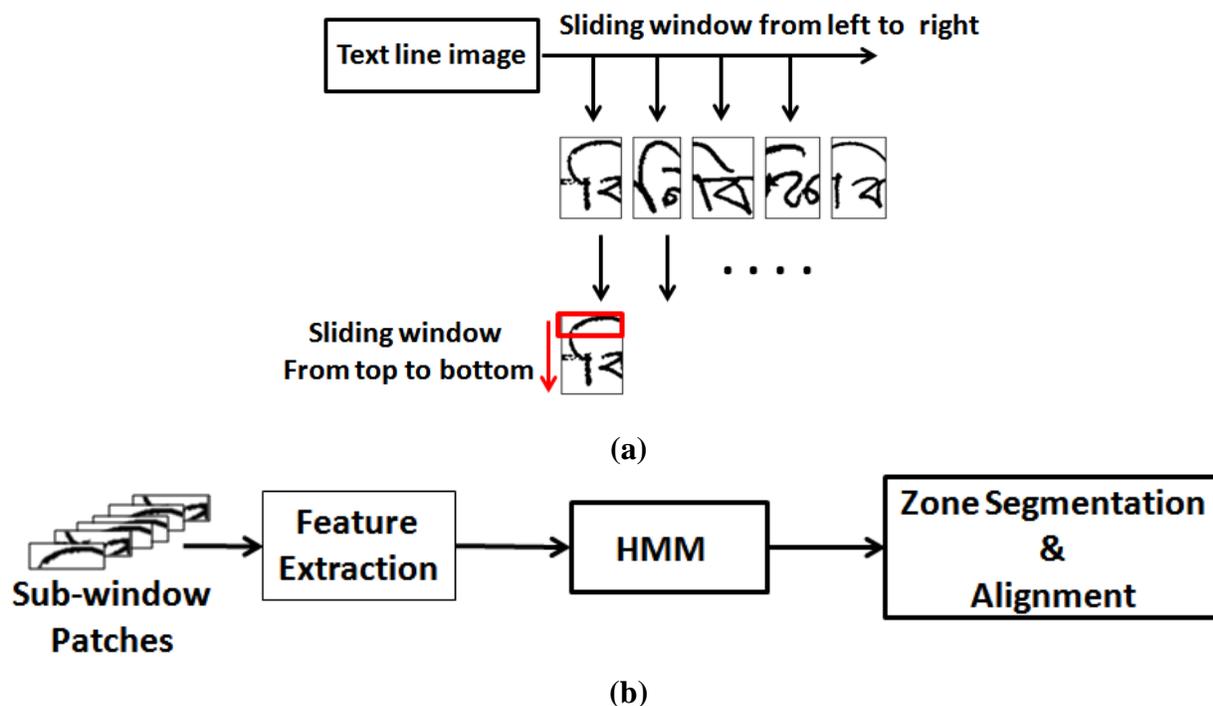

**Fig.9: The framework of HMM-based zone segmentation. (a) Patch extraction from a line image using sliding window. (b) HMM is used to identify the zones in each position.**

### 3.1. Zone segmentation using HMM

Each word image from the training dataset is segmented to build the HMM model of upper, middle and lower zone components. The procedure is as follows.

We estimated the height ($H_L$) of the text line using a height histogram computed from water reservoir heights. In handwritten text lines, there exists a variety of writing styles. So, simple connected component analysis cannot provide text line height information. To get proper height information we take the mode of Bottom reservoir height [6]. Next, the text line is processed using HMM for zone segmentation. For this purpose, a sliding window is first chosen to slide horizontally across the line image. The image patch that moves from left to right is of width ($W_P$) where $W_P = \alpha \times H_L$. From experiment, $\alpha$ is set as 1.5. Next, in each position of this window, a sub-window of height ($H_P$) and width $W_P$ is moved vertically from top to bottom. The length of $H_P$ is chosen experimentally. Note that, the sliding sub-window (see Fig.9.) may not contain the entire modifier (upper/lower zone modifier) in each position. The sub-window at each position collects the patches and corresponding features. Next, these features are training using HMM for zone-wise component modeling. PHOG features are extracted from each patch for zone segmentation.

**3.1.1. Zone Separation and Alignment:** The image patches from a text line image are extracted and next these patches are segmented into upper, middle or lower zones using HMM. The image patches are trained using a zone-component filler model (see Fig.10.) to restrict the zone identification results of the modifiers corresponding to Indic script characteristics. Next, Viterbi algorithm has been used to get the alignments of the zone-components. With this approach, two zone segmentation lines, upper and lower alignment lines can be obtained for each image patch (see Fig.11(a(II)) or Fig.11(b(II))).The alignment lines (i.e. the segmentation lines) of upper and lower zones may not be uniform in each horizontal sliding window. To make these boundary lines of zone segmentation uniform, the mid-point of alignment lines at each position are joined by a straight line (see Fig. 11(a.III) or Fig.11(b.III)).

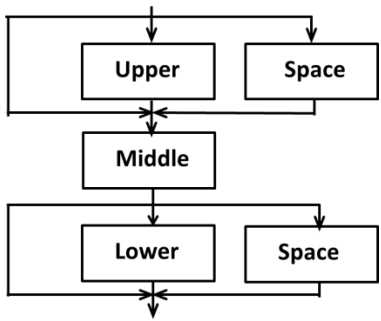

**Fig.10: Filler model used for HMM-based zone segmentation.**

**3.1.2. Segmentation Improvement with Context Information:** Sometimes, the alignment lines at some subsequent image patches at almost same position (see the upper zone-line of 3$^{rd}$ and 4$^{th}$ image patch in Fig.11.(a.II)). Also, it may happen that due to complex shape of modifiers, the alignments of upper and lower boundary are not proper. There, it was noted that HMM fails to recognize the correct sequence for upper, middle and lower portion of a word image for those particular cases. In Fig.12.(a) it has been shown, that particular image patch is being recognized as <Space><Middle><Space> instead of <Upper><Middle><Space>. To take care of these problems, context information of the alignment is considered. The position of each zone is refined by taking average of the preceding and following patches (see Fig.12.(b)).

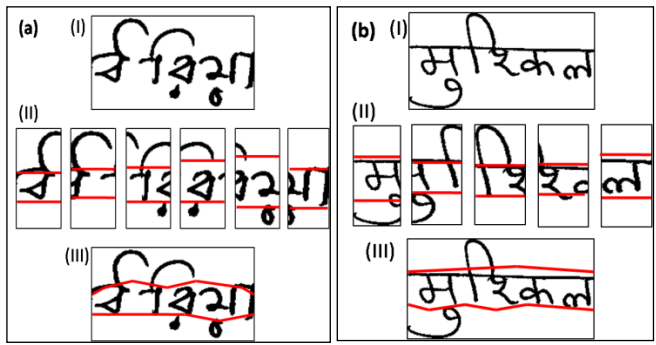

**Fig.11: Alignment extraction from image patches and approximating zone segmentation boundary**

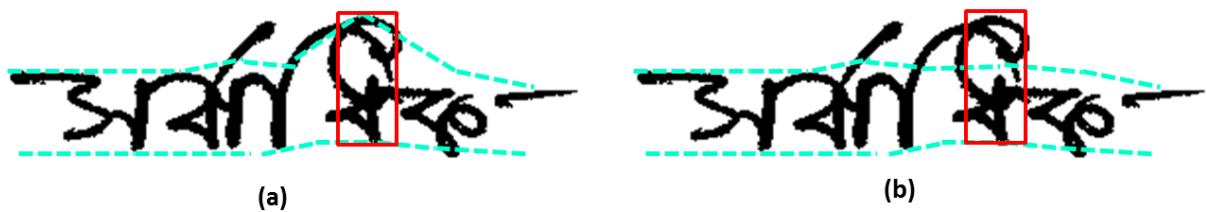

**Fig.12. Example showing improvement in zone segmentation using context information in (b) in comparison with (a)**

**3.2. Word Spotting using Middle-Zone**

Zone segmentation plays an important role in reducing the character classes of Bangla and Devanagari script for HMM's character training model. At first, the two trajectories, those separate the middle zone from upper zone and lower zone are detected using HMM based zone segmentation. Portions between these two trajectory lines, considered as middle zone of the text line image. This portion contains the major information of text word images.

Simultaneously, taking care of the fact that combination of foreground and background information improves the word spotting efficiency, both of these information were considered in middle zone based word spotting. To obtain the background information of the middle zone, the zone between upper and lower trajectory paths of are used as mask in the original image of text line image after forming the bottom reservoir [11]. Then particular portions of the reservoir are extracted which falls between these two trajectory lines. Then background features are extracted and get concatenated to foreground feature. Above process has been demonstrated in Fig.13, diagrammatically.

As we have adopted the approach of zone segmentation for our word spotting framework, it demands a mapping function of Indic keywords while searching. The ASCII keywords given by user are mapped to middle-zone based keywords. To do so, each character from full word level is mapped to middle zone level by a function according to a set of rules, eg. the character modifiers "□□" and "□□" will be "□□" and "□" respectively. In Table I, it has been shown, some examples of full words are being converted to their respective middle zone which are used as query word for word spotting in our zone segmentation based approach.

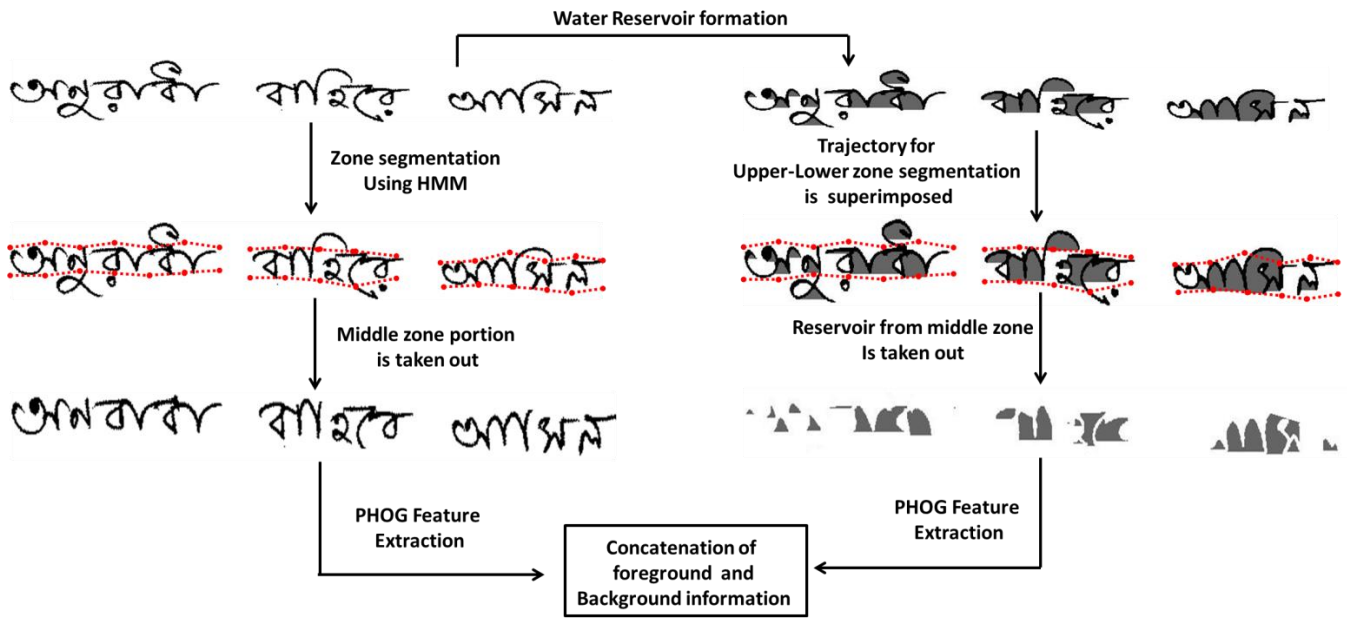

**Fig.13. Middle zone based word spotting using foreground and background information.**

**Table I: Mapping of keywords from full word to middle zone**

| Bangla | | Devanagari | |
|---|---|---|---|
| Full word | Middle zone | Full Word | Middle zone |
| সুকীর্তির | সকাতব | भुमिका | भामका |
| অনুদিত | অনাদত | लेकिन | लाकन |
| সাইনবোর্ড | সাহনবোড | पुलिस | पालस |

### 3.3. Combination of Zone based Retrieval

Middle zone segmentation is very much effective to train the HMM character model because of significant reduction in the number of character sets. Zone-wise segmentation approach increases the word spotting performance for Indic script significantly. But HMM based zone segmentation is not always accurate. It may fail sometimes due to skew nature of word, complex writing with missing "Matra", etc., which may lead to losing of information from text line image. Also, due to reduced content from middle zone, the text component from middle zone may match with other words. For example, different words like '☐☐☐, '☐☐☐', '☐☐☐'will be reduced to '☐☐☐' using middle zone portion. Because, the middle zone portion of this word image is segmented, the distinguishing features from upper and lower zones are neglected. So, searching with similar middle-zone mapped keywords will provide false positives which need special care while searching. Other zone-wise information will be useful to overcome this problem.

Hence, combination of word-spotting performance using zone-based information was used in our system to overcome the shortcoming of middle-zone based spotting system. By combining zone information, the zone-wise information will complement each other for word retrieval. The combined system will be used to re-ranking the retrieval result obtained from middle zone based approaches. To combine the zone-wise information, we first detect the word location by middle-zone based spotting in each line. It is performed using Viterbi alignment of keyword model. Once, the location of the keyword is identified we search the upper and lower zone modifiers in that word position. For this purpose, we compute the upper (lower) profile of the upper (lower) zone of corresponding location. The number of peak in upper/lower profile is used to reject the false positive results from word spotting results. We count the number of peaks in upper profile as well as lower profile. To avoid error due to noise, a Run Length Smoothing Algorithm (RLSA) [6] was used in upper and lower zone before profile computation. To count the number of upper-lower modifiers, we look at the depth of the valley between two consecutive peaks (See the peak 1 in lower modifier zone in Fig.14(a)). If the depth of the valley between two consecutive peaks reaches the middle zone profile, we separate them as two different modifiers, otherwise it is marked as a single modifier. For a particular keyword, the number of upper-lower modifiers determination procedure is shown graphically in Fig.14. Fig.15. explains the combination of zone information to re-rank the word spotting results.

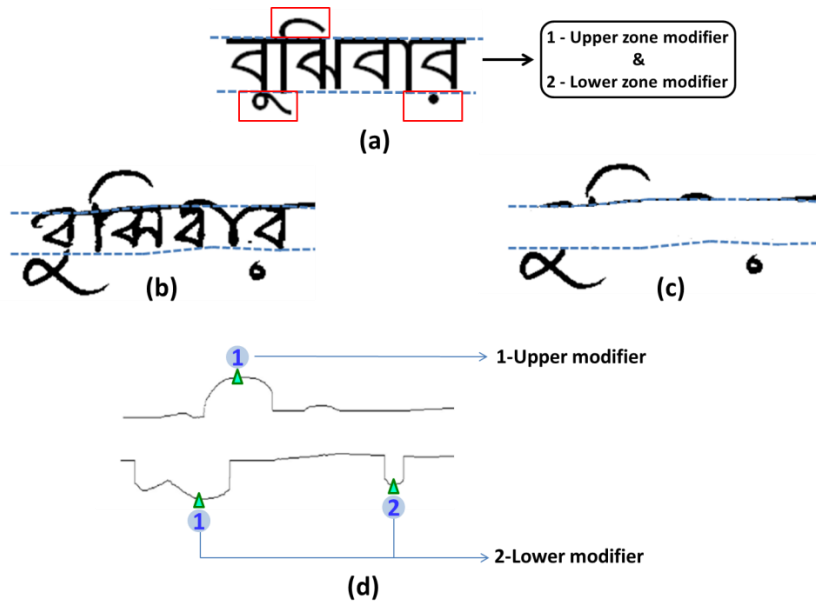

**Fig.14: (a)'□□□□□□' is a Bangla word with 1-Upper and 2-Lower modifiers. (b) Trajectory lines segmenting the word in three zones are found using HMM (c) Upper-Lower zone portion is taken out (d) No of modifiers is calculated using the profile information of upper-lower zone.**

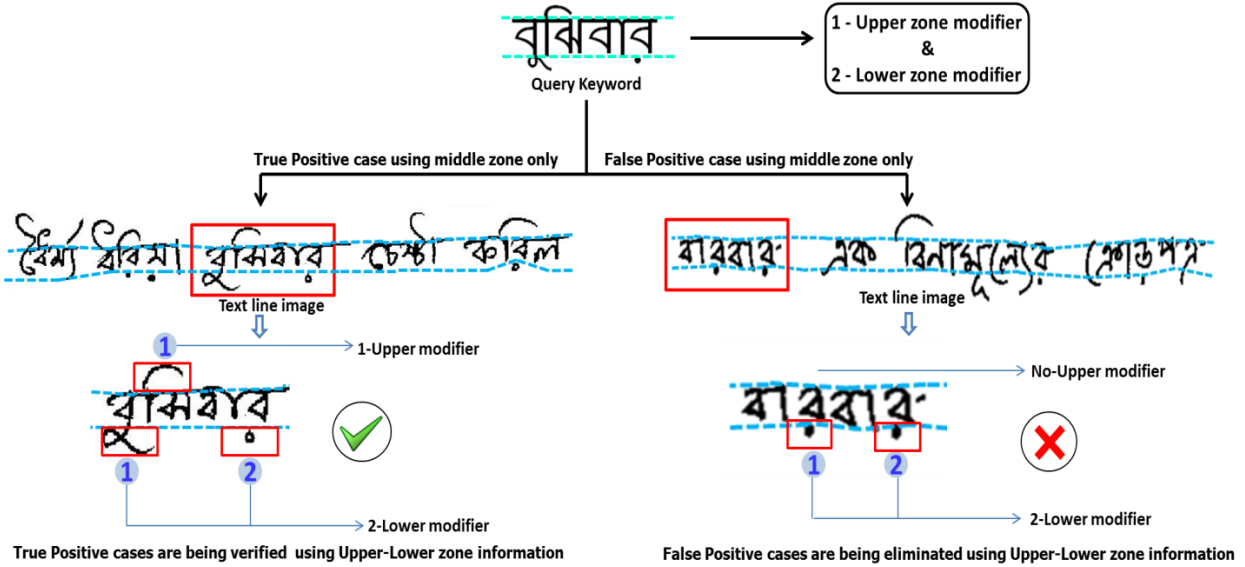

**Fig.15: Re-ranking of results using Upper-lower zone information.**

---

**Algorithm 1.** Zone information combination for full word spotting

**Require:** Middle zone based word spotting using keyword ($K_w$) in text line
**Ensure:** Re-ranking of word-spotting results

**for** each text-line $T_i$ of $T_1...T_N$ **do**

Step 1: From middle-zone based word-spotting information, find start ($L_s$) and end ($L_f$) locations of $K_w$ in $T_i$ using Forced Viterbi alignment of keyword model.

Step 2: Extract the upper ($Z_u$) and lower ($Z_l$) zone image from $L_s$ to $L_f$ in $T_i$

Step 3: Compute upper profile in $Z_u$ and calculate the number of peaks ($N_u$) in upper profile.

Step 4: Compute lower profile in $Z_l$ and calculate the number of peaks ($N_l$) in lower profile.

Step 5: From ASCII text of $K_w$, find the number of upper ($M_u$) and lower ($M_l$) modifiers.

Step 6: **If** (($N_u != M_u$) || ($N_u != M_u$))  Eliminate $T_i$ as it generated false positive.

//text line with false positive result

**end if**

**end for**

## 4. Experiment Results & Discussions

### 4.1. Data Details

We have collected document images written by different professional. The input of our system can be either arbitrary keyword string or text line image. We have collected word images of different writer for both Bangla and Devanagari script. Then we have generated a total of 8592 line images for Bangla and 7902 line images for Devanagari; both containing two to eight word images in a line. To test the effectiveness of background information, we have also considered IAM (English) dataset. The details of data used in our experiment are shown in Table II. Some of the keywords used for word spotting in 3 scripts are shown in Table III.

We have measured the performance of our word spotting system using precision, recall and mean average precision (MAP). The precision and recall are defined as follows.

$$Precision = \frac{TP}{TP + FN} \quad Recall = \frac{TP}{TP + FP}$$

Where, TP is true positive, FN is false negative and FP is false positive. MAP value is evaluated by the area under the curve of recall and precision.

**Table II: The dataset of text lines and keywords used for the experiment**

|  | Bangla | Devanagari | IAM(English) |
|---|---|---|---|
| **Training** | 6824 | 6214 | 6161 |
| **Validation** | 854 | 810 | 920 |
| **Testing** | 914 | 878 | 929 |
| **Keywords** | 612 | 623 | 882 |

**Table III: Some examples of keywords**

| Bangla | Devanagari | IAM (English) |
|---|---|---|
| লাগাতার | खिलाफ | being |
| ফেরিওয়ালা | परियोजना | House |
| মতপার্থক্য | भेजने | Government |
| মহানগরী | शिकायत | People |
| শশীকর | अधिकतर | would |

| সংবাদদাতা | গোলাবাড়ী | should |

## 4.2. Word Spotting Performance with Foreground and Background Information

For our experiment, 32 Gaussian mixture and 6 states provided optimum results. We have evaluated the performance for word spotting considering local threshold and global threshold both in the Fig. 16. For local threshold single image has been considered for optimization of the threshold value whereas a standard value has been used for all query keyword in case of global threshold. We have considered a total of 612 and 623 keywords for our word spotting performance evaluation in Bangla and Devanagari scripts respectively.

A comparative evaluation is shown in Fig.17 for combination of foreground and background information with foreground or background information alone. It has been observed that there is a significant improvement in the word spotting performance using our combined feature extraction method.

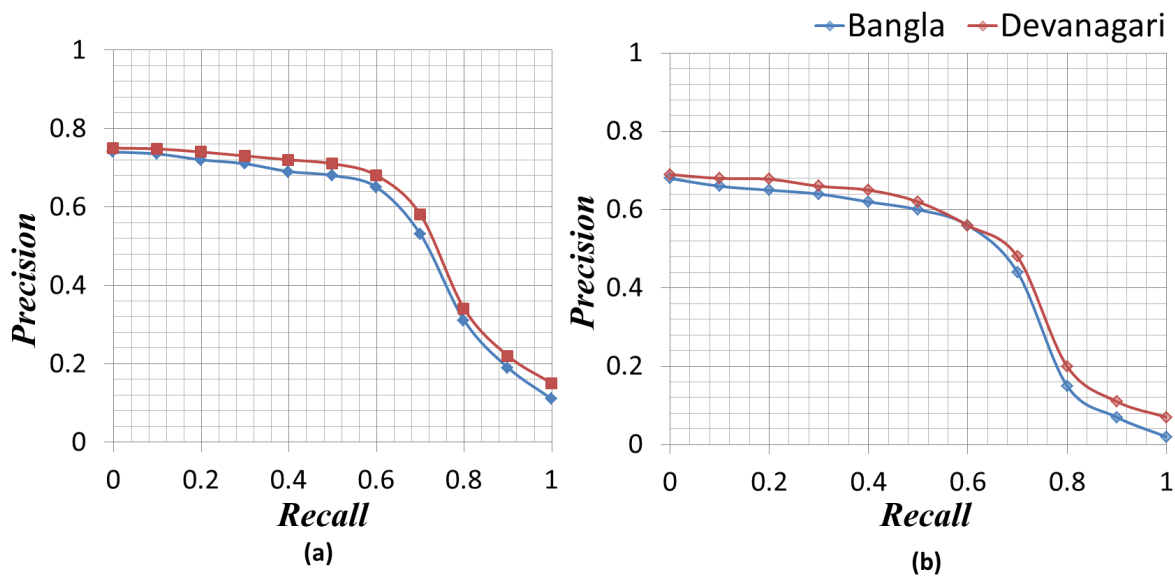

**Fig.16: Word Spotting Performance taking (a) Local threshold and (b) global threshold (using foreground information only).**

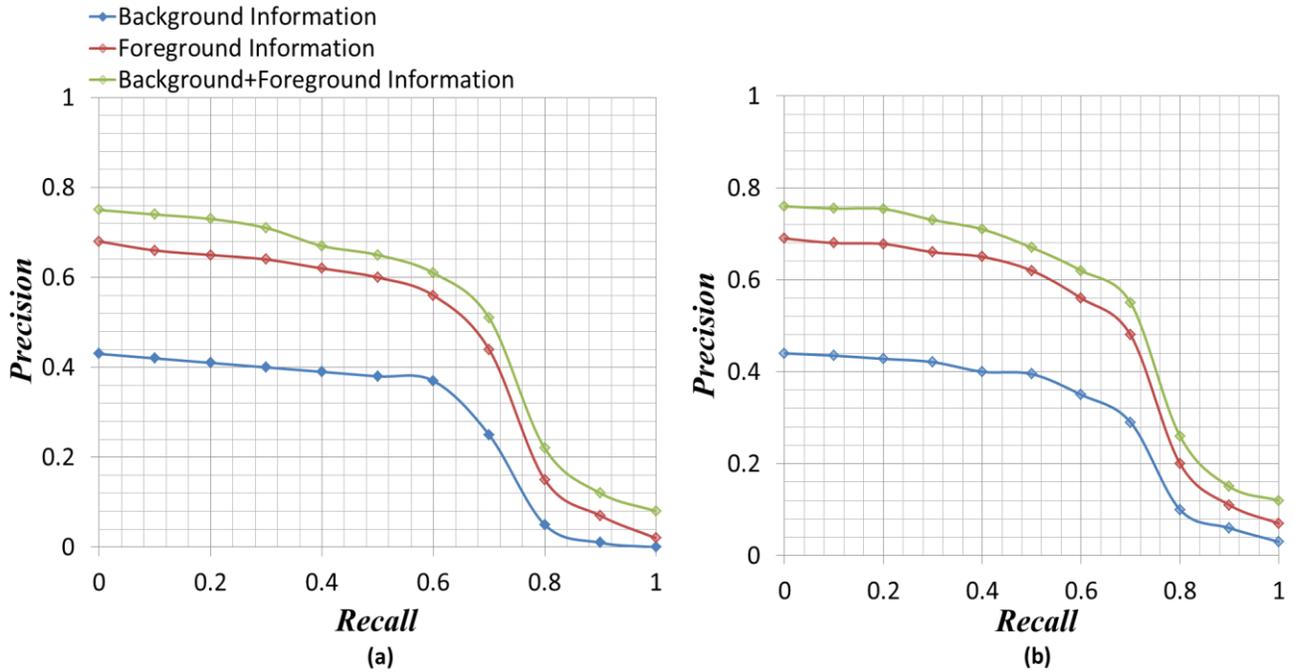

**Fig.17: Comparative study of word Spotting performance on (a) Bangla (b) Devanagari and using concatenation of foreground and background information with foreground or background information alone using global threshold.**

Also we have checked the result using different number of keywords in our dataset considering global threshold using concatenation of foreground and background features. The results are shown in Fig.18 based on the precision recall curve.

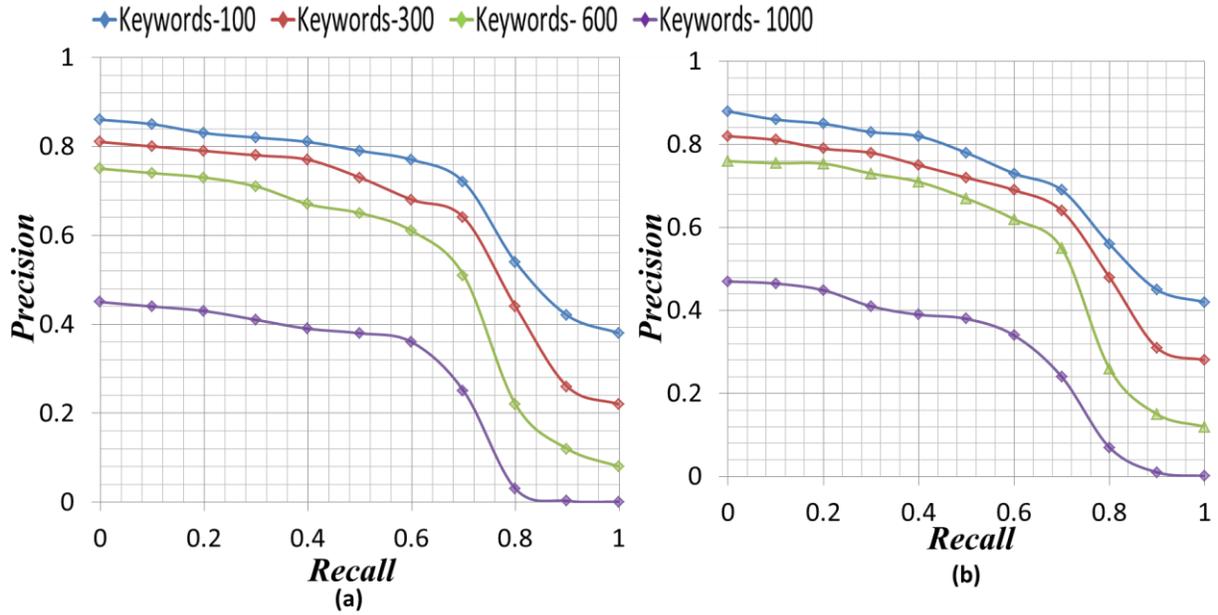

**Fig.18: Comparative study of word spotting performance on (a) Bangla (b) Devanagari script with different number of keywords**

A comparative study of 2 different features, namely LGH and PHOG, is also performed to check the efficiency of our feature extraction approach by concatenating foreground and background information. LGH feature [15] which found to be useful in Latin text also gives a close accuracy to PHOG feature. In LGH features, with 8 bins we found 128 dimensional feature vector for each sliding window position. Word spotting performance is slightly found to be better using PHOG feature than LGH feature. MAP value has been given in Table IV.

**Table IV: Comparison of LGH and PHOG feature in word spotting using global threshold**

| Script | LGH | PHOG |
|---|---|---|
| Bangla | 51.84 | 52.64 |
| Devanagari | 52.55 | 53.71 |
| IAM (English) | 48.04 | 48.98 |

| Query keyword | Text line Image | (i) | (ii) |
|---|---|---|---|
| সভাপতি | কলকাতায় প্রদেশ কংগ্রেস সভাপতি | ✗ | ✓ |
| পথচারীর | খুঁচিয়াত এ পথচারীর অথা লইয়া | ✗ | ✓ |
| दुनिया | जीतने वाला दुनिया का तीसरा | ✗ | ✓ |
| शिकायत | लिखित शिकायत दी हूँ जिसे | ✗ | ✓ |

**Fig.19: Examples showing qualitative word spotting performance using (i) foreground information only (ii) combination of foreground and background information where correct and incorrect results are indicated by tick and cross marks respectively.**

**Experiment in IAM dataset:** To evaluate the performance of background feature, we have used IAM dataset for comparative studies. The test dataset of IAM English sentence dataset [34] has been used to evaluate the performance. Total 929 text lines and 882 keywords were considered for comparison. Next, foreground, background information and combination of foreground and background features were extracted from these text images and HMM based word spotting is applied for word spotting. To summarize the performance of a system, the average precisions are measured for the 821 keywords and the mean is reported. Table V shows the MAP with best performance obtained using combined features.

**Table V: Analysis in IAM dataset for English**

| Features | MAP (Local) | MAP (Global) |
|---|---|---|
| Foreground Information | 69.58 | 48.98 |
| Background Information | 44.28 | 32.19 |
| Foreground + Background | 72.28 | 51.87 |

## 4.3. Zone Segmentation Analysis

Next, zone segmentation is performed in text line images to perform word spotting using middle zone. The zone-segmentation approach proposed in Section 4.1 has been applied to obtain the middle zone portion. PHOG feature has been used for feature extraction. For experiments, we evaluated our zone segmentation algorithm on a dataset of 7,128 Bangla and 6,974 Devanagari word images. We collected 7267 and 5184 image patches from Bangla and Devanagari dataset respectively and labeled them manually. We have used continuous density HMMs with diagonal covariance matrices. For the experiment we considered 32 as Gaussian mixture and 8 as state number for HMMs. For a qualitative comparison, some examples of zone segmentation using the proposed zone segmentation method and the previously proposed projection analysis based method are shown in Fig.20. It can be noted from the table that the learning based method has outperformed the projection analysis based method. For more clarification, quantitative results are also shown later.

**4.3.1. Zone Segmentation using Projection Analysis**

Here we have compared HMM based zone segmentation method with classical projection based method. Projection based zone segmentation is as follows: In Bangla and Devanagari script 'Matra' is present which separates the upper zone modifier from middle zone portion. Matra is determined by projection analysis in horizontal direction and considering the row with highest peak. Simultaneously, to separate the lower zone modifiers from the middle zone, horizontal projection profile is considered generally. The pixels in the middle zone is more cluttered and closed, where as pixels in the lower zone are sparsely distributed. Sharp decline in projection peak is observed when it moves from middle zone to lower zone. That particular row which shows a sharp decline in histogram, is considered to be the demarcation line between middle zone and lower zone. Projection based method is being applied globally on a text line image and also locally on successive parts of a line image in a sequential manner.

For global purpose, we segment the upper-lower zone modifiers using projection profile of the whole text line image. This gives highly imprecise result because of free flow nature of handwriting makes a handwritten text line image not to remain in a perfect straight line. For this we extend this global projection analysis method to local projection analysis to get better zone segmentation of a text line image. Here we have taken out small portion of a text line image and find out the upper-lower zone trajectories for each text line segment and an average estimation is done to merge the successive upper-lower zone trajectories of consecutive text line image segments.

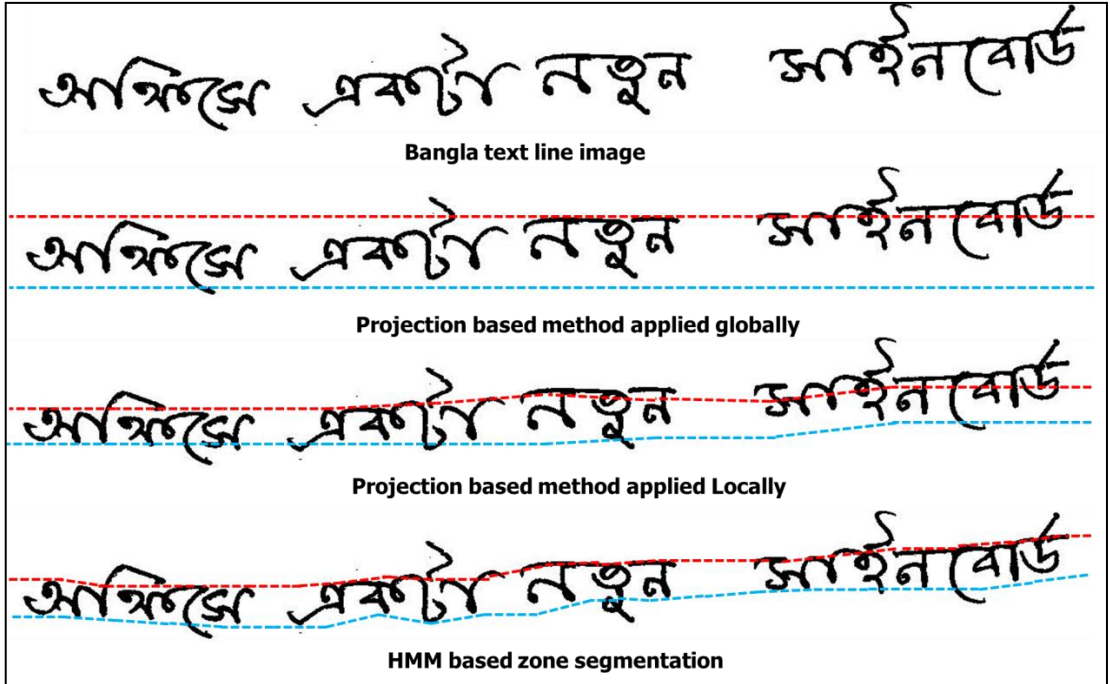

**Fig. 20: Qualitative results of zone segmentation using Projection analysis and HMM based alignment**

## 4.4. Word Spotting performance using Middle Zone

After segmenting the text lines into different zones by HMM, we use middle zone portion for word spotting experiment. As discussed in Section 4.2, we have used foreground information, and combining background information for middle zone based word spotting experiment. Fig.21 shows the comparative studies of Indic word spotting performance using these 2 different approaches qualitatively.

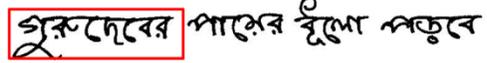

**Fig. 21: Example showing qualitative word spotting performance using (i) without zone segmentation - i.e. in text line images of all zones (ii) with zone segmentation, here, query Keyword is first mapped to its equivalent middle zone components and word spotting is done in zone segmented text line images using that middle zone equivalent query keyword. This diagram gives few examples where zone segmentation based word spotting approach has been able to spot the query keyword perfectly but full zone based method fails.**

We have tested the middle zone-based word spotting with zone segmentation using HMM and projection analysis. The global projection based segmentation did not work in many text lines because of slant nature in text lines. To avoid it, text lines are divided into strips and local projection analysis was performed into each strip for zone segmentation. The division of strips were tested with different length and the best performance was found with length 1/10th of the line width. The comparative studies in Bangla and Devanagari scripts are shown in Fig.22.

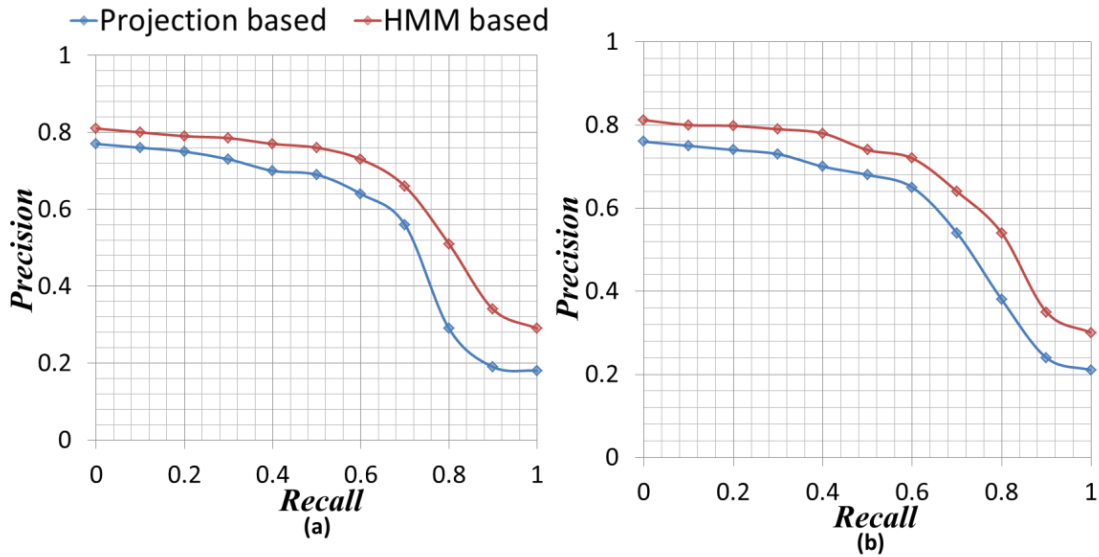

**Fig.22: Comparison of word spotting performance using HMM-based alignment and projection based segmentation using middle zone only.**

## 4.5. Combination of Zone-based Information

After word spotting using middle zone portion of text line image, we include the information regarding the total number of upper-lower zone modifiers for that particular word. This improves the precision of word spotting in both Bangla and Devanagari scripts. In Fig. 23, Precision – Recall curve for both Bangla and Devanagari scripts are given when upper-lower zone information is combined with middle zone result.

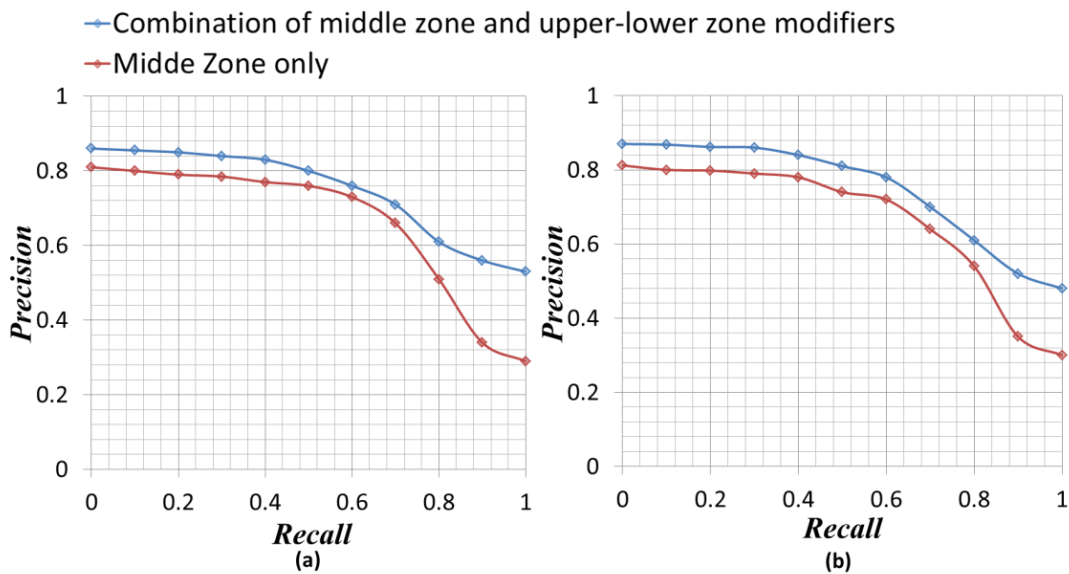

**Fig.23: Comparison of word spotting performance using middle zone only and modifiers combination for (a) Bangla and (b) Devanagari scripts.**

**Table VI. Global MAP values using different method**

| Approach | Feature Type | Bangla | Devanagari |
|---|---|---|---|
| Without using zone segmentation | Foreground | 49.56 | 50.17 |
|  | Foreground + Background | 52.64 | 53.71 |
| Using HMM based zone segmentation (middle zone only) | Foreground | 62.64 | 63.58 |
|  | Foreground + Background | 65.84 | 66.67 |
| Combination of middle zone and upper-lower zone modifiers | Foreground + Background | 72.62 | 73.12 |

The scalability of our system is tested with increasing number of keywords in both Bangla and Devanagari word spotting. Fig.24 shows the precision-recall results with 100, 300, 600 and 1000 keywords. With increasing number of keywords the performance drops. Keeping recall to 0.6, the precision is higher than 75% with 600 keywords. We have also checked the performances using keyword of different lengths considering global threshold which are shown in Fig.25.

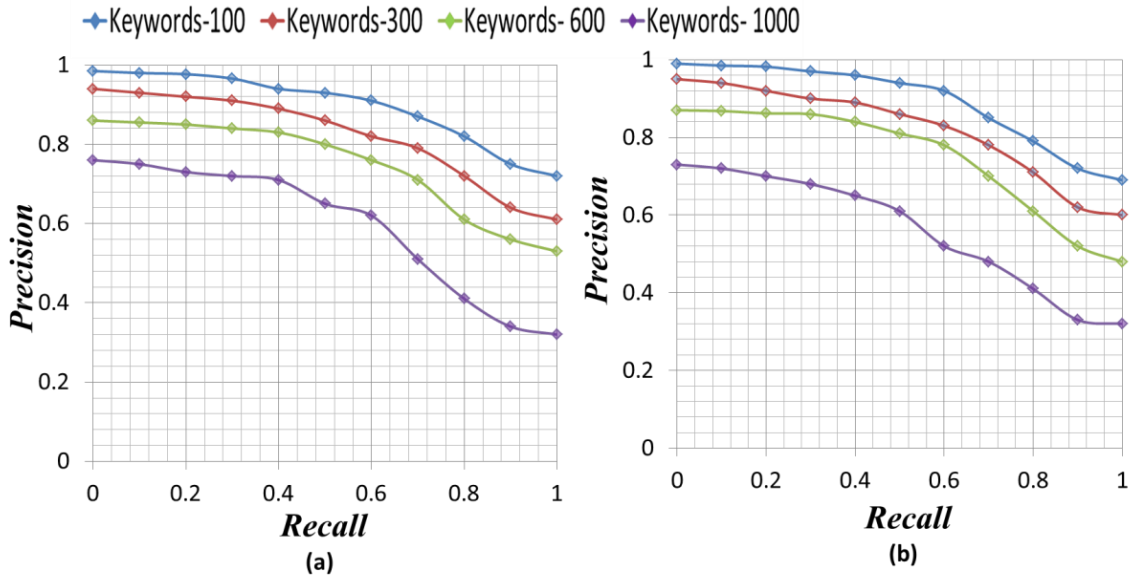

**Fig. 24: Comparative study of word spotting performance on (a) Bangla (b) Devanagari script with different number of keywords**

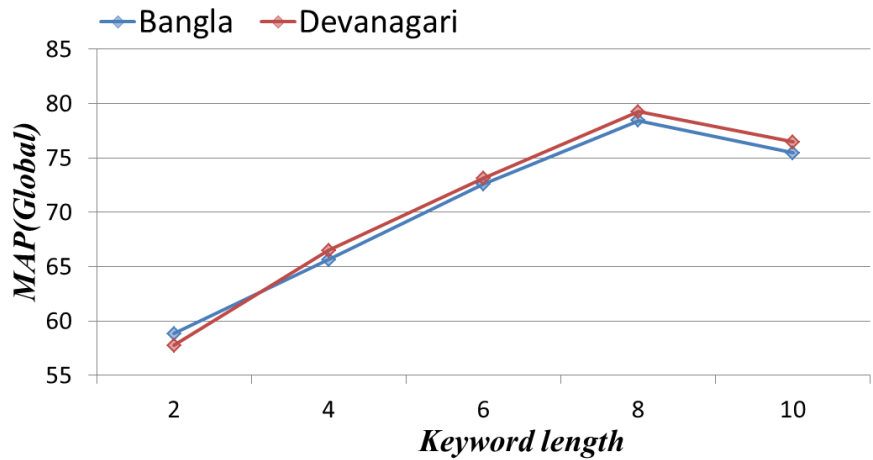

**Fig. 25: Word spotting performance using keywords of different length**

## 4.6. Comparison with Other Methods

We have compared our zone-based word spotting performance with the traditional full word wise spotting performance. Fig.26 shows the comparative studies of foreground-information based word spotting using zone-wise and without-zone segmentation based approaches. In full word-wise spotting performance, sliding window features extracted using PHOG was fed into HMM for line scoring. We noticed that when the background information is combined with foreground, the performance of zone-wise spotting outperforms without zone based approach and only foreground information-wise approach (See Fig. 26).

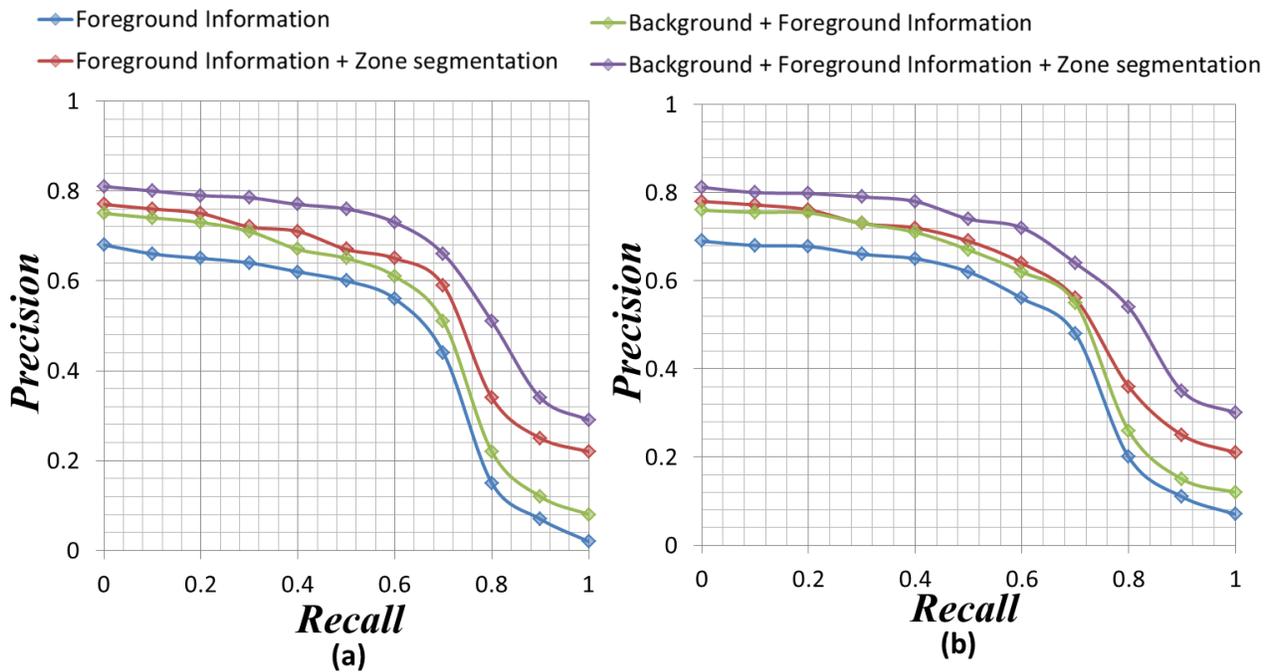

**Fig. 26: Comparative study of word spotting performance on (a) Bangla (b) Devanagari with different methods using global threshold.**

**Comparison with DTW:** To compare our approach using traditional word matching method, DTW-based sequence matching [3] is used for similarity measure of handwritten words. The sequences are ''warped'' non-linearly in the time dimension to determine a measure of their similarity, independent of certain non-linear variations in the time dimension. A word is represented as 4 sequences, which are computed from four profiles features of a word. Four components such as vertical projection profile, upper profile, lower profile and vertical crossing (vertical ink transition) have been extracted as discussed in [18] for word-level features. The value of all the profile features is normalised to the range [0–1]. The DTW-based technique for measuring similarity between two sequences uses the Sakoe-Chiba band [30] to speed up the computation. The performance using DTW was found not satisfactory. We have obtained precision of 18.62% and recall of 20.23% respectively.

### 4.7. Parameter Evaluation

We considered continuous density HMMs with diagonal covariance matrices of GMMs in each state. A number of Gaussian mixtures were tested on validation data. Word spotting performance with different Gaussian numbers is detailed in Fig.27. It was observed that with 32 Gaussian Mixture, all features provide the best results. The best state number was found to be 6. Fig.27 illustrates the performance with varying the state number on word spotting experiment. In word spotting the PHOG features are extracted from text line image with a sliding window of height 40 and width 6. The window slides with 50% overlapping in each position. Different values of width were tested but with size 40x6 we obtained best results. During zone segmentation purpose, the sub-window patch was considered as 40x8. We also tested with different height and width, but the sliding window size with 40x8, we obtained the best results.

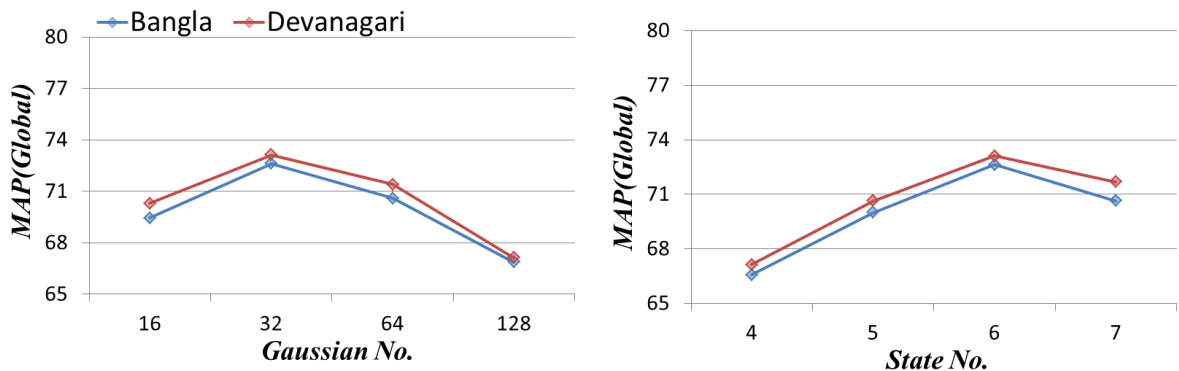

**Fig. 27: Word spotting performance evaluation using different Gaussian number and State number.**

### 4.8. Time computation

Experiments have been done on an I5 CPU of 2.80 GHz and 4G RAM664 Computer. For each query, the average runtime has been computed from different runs made in the experiment. In Table 1 we show the time taken using queries of various approaches developed in MATLAB. The average time required for both Bangla and Devanagari to give the result for the occurrence of a given keyword in a particular text line is given in Table.VII.

**Table VII: Time computation analysis of word spotting using different approaches.**

| Approach | Feature Type | Time |
|---|---|---|
| Without using zone segmentation | Foreground | 0.76 Sec |
| | Foreground + Background | 1.35 Sec |
| Using zone segmentation (middle zone only) | Foreground | 1.44 Sec |
| | Foreground + Background | 1.86 Sec |
| Combination of middle zone and upper-lower zone modifiers | Foreground + Background | 2.14 Sec |

## 5. Conclusion and future work

In this paper we have proposed a novel feature extraction method combining foreground and background features for word spotting. Line level word spotting provides better performance than word segmenting approach where the gap between two consecutive words is not regular. The proposed system separates a text line into 3 zones and use word spotting approach in middle zone. This zone segmentation based approach leads to reduction in the number of character sets in Devanagari and Bangla scripts. A novel learning based zone segmentation method is proposed for text line images which works well even in case of any curved or skewed text line image. This learning based zone segmentation method out performs the traditional projection based zone segmentation method to a large extent in our dataset. As the proposed approach segments the zones of words, the training data entailed for character modelling is less and the zone-based word spotting outperforms traditional word spotting approaches. We noted that PHOG feature outperformed the LGH feature for word spotting performance. Finally, zone-wise results are combined together to improve the word spotting performance.

## 6. Conflict of Interest

None.